\documentclass[10pt,twocolumn,letterpaper]{article}

\usepackage{wacv}
\usepackage{times}
\usepackage{epsfig}
\usepackage{graphicx}
\usepackage{amsmath}
\usepackage{amssymb}
\usepackage{booktabs}
\usepackage{pifont}
\newcommand{\cmark}{\ding{51}}%
\newcommand{\xmark}{\ding{55}}%
\usepackage{multirow, makecell}
\usepackage{placeins}
\usepackage{caption}
\usepackage{subcaption}
\usepackage{comment}
\usepackage{booktabs}
\usepackage[table, svgnames, dvipsnames]{xcolor}
\usepackage{arydshln}

\wacvfinalcopy

\ifwacvfinal
\usepackage[breaklinks=true,bookmarks=false]{hyperref}
\else
\usepackage[pagebackref=true,breaklinks=true,colorlinks,bookmarks=false]{hyperref}
\fi

\begin{document}

\title{Masked Autoencoder for Self-Supervised Pre-training on Lidar Point Clouds}

\author{Georg Hess$^{1,2}$
\quad
Johan Jaxing$^{1}$
\quad
Elias Svensson$^{1}$
\quad 
David Hagerman$^{1}$
\\
Christoffer Petersson$^{1,2}$
\quad
Lennart Svensson$^{1}$
\\
\normalsize$^1$Chalmers University of Technology \hspace{1cm} $^2$Zenseact\\
{\tt\small georghe@chalmers.se}
}

\maketitle

\begin{abstract}
    Masked autoencoding has become a successful pretraining paradigm for Transformer models for text, images, and, recently, point clouds.
    Raw automotive datasets are suitable candidates for self-supervised pre-training as they generally are cheap to collect compared to annotations for tasks like 3D object detection (OD).
    However, the development of masked autoencoders for point clouds has focused solely on synthetic and indoor data. Consequently, existing methods have tailored their representations and models toward small and dense point clouds with homogeneous point densities.
    In this work, we study masked autoencoding for point clouds in an automotive setting, which are sparse and for which the point density can vary drastically among objects in the same scene.
    To this end, we propose Voxel-MAE, a simple masked autoencoding pre-training scheme designed for voxel representations. We pre-train the backbone of a Transformer-based 3D object detector to reconstruct masked voxels and to distinguish between empty and non-empty voxels.
    Our method improves the 3D OD performance by 1.75 mAP points and 1.05 NDS on the challenging nuScenes dataset. Further, we show that by pre-training with Voxel-MAE, we require only 40\% of the annotated data to outperform a randomly initialized equivalent. 
\end{abstract}

\section{Introduction} 
Self-supervised learning enables the extraction of rich features from data without the need for human annotations. This has opened up new avenues where models can be trained on ever-larger datasets. Fueled by robust representations, self-supervised models have seen great success in fields such as Natural Language Processing (NLP) \cite{brown2020language,devlin-etal-2019-bert,radford2018improving} and computer vision (CV) \cite{caron2021emerging,chen2020simple,he2021masked}. Specifically, masked language modeling  \cite{devlin-etal-2019-bert} and masked image modeling \cite{bao2022beit,he2021masked,zhou2021ibot} have proven themselves as simple, yet effective, pre-training strategies. Both of these approaches train models to reconstruct sentences, or images, from partially masked inputs. Subsequently, models can be fine-tuned toward downstream tasks, often outperforming their fully supervised equivalents.

\begin{figure}[t]
    \centering
    \includegraphics[width=0.9\linewidth,trim={0cm 6.5cm 15.6cm 0cm}, clip,page=5]{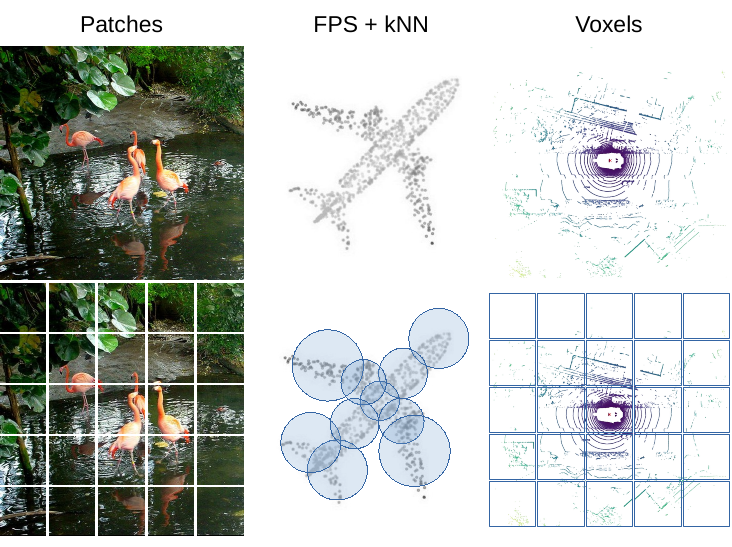}
    \caption{MAE \cite{he2021masked} (left) divides images into non-overlapping patches of fixed size. Existing methods (middle) for masked point modeling create point cloud patches with a fixed number of points by using furthest point sampling and k-nearest neighbors. Our method (right) uses non-overlapping voxels with a \textit{dynamic} number of points. Airplane point cloud from \cite{pang2022masked}. 
    }
    \label{fig:patch_fpsknn_voxel}
\end{figure}

Autonomous driving is an application well-suited for self-supervised pre-training strategies, including masked autoencoding. 
In the automotive domain, the collection of raw data is relatively cheap, while annotations for common tasks such as object detection (OD), tracking, and semantic segmentation are expensive and time-consuming to acquire. 
Especially for data in 3D, the sparsity of lidar and radar sensors can make labeling labor-intensive and even ambiguous. 
Self-supervised pre-training is thus an appealing alternative to create robust and general feature representations, and ultimately reduce the need for human-annotated data. 

Recently, multiple works have applied masked point modeling techniques to pre-train point cloud encoders \cite{fu2022pos,liu2022masked,pang2022masked,Yu_2022_CVPR,zhang2022point}. 
These have achieved favorable results on downstream tasks like shape classification, shape segmentation, few-shot classification, and indoor 3D OD, indicating the effectiveness of masked autoencoders in the point cloud domain. 
However, evaluation has been focused on synthetic data such as ShapeNet \cite{shapenet2015} and ModelNet40 \cite{wu20153d}, and indoor datasets like ScanObjectNN \cite{uy2019revisiting}, ScanNet \cite{dai2017scannet}, and SUN RGB-D \cite{song2015sun}. 
Compared to automotive point clouds, these datasets contain many points for all objects and the point density is generally constant within a scan, making the detection and classification of objects less challenging. 

Further, existing methods have tailored design choices like point cloud representation and model selection to dataset characteristics. 
For instance, fewer points per scene lessen requirements on computational efficiency and enable the use of vanilla Transformers \cite{liu2022masked,pang2022masked,Yu_2022_CVPR}. 
Moreover, previous works rely exclusively on furthest point sampling (FPS) and k-nearest neighbors (kNN) for dividing point clouds into subsets of equally many points, see Fig.~\ref{fig:patch_fpsknn_voxel}. 
This works well when point clouds are evenly distributed and simplifies the reconstruction during pre-training, as the model predicts a fixed number of points for each subset. 
However, this representation is sub-optimal for efficiently solving downstream tasks in the automotive domain. 
First, there is a risk of discarding points, as shown at the wing tips in Fig. \ref{fig:patch_fpsknn_voxel}. 
This potential loss of information makes it ill-suited for safety-critical applications. 
Second, the representation is redundant as subsets may overlap, creating unnecessary computational load.

In this work, we propose to use masked point modeling in an automotive setting. 
To this end, we present Voxel-MAE, a masked autoencoder pre-training strategy for \textit{voxelized} point clouds, and deploy it on the large-scale automotive dataset nuScenes \cite{caesar2020nuscenes} to study its effects on 3D OD. 
The voxel representation is widely used in 3D OD due to its ability to efficiently describe large point clouds but has not been used previously for masked autoencoder pre-training. 
To capture the unique nature of voxels during reconstruction, we propose a unique set of loss functions to capture shapes, point density, and the absence of points simultaneously. 
In comparison to previous approaches, such as Point-BERT \cite{Yu_2022_CVPR} and POS-BERT \cite{fu2022pos}, our method is simpler in the sense that it does not rely on training a separate tokenizer for embedding and reconstructing the point cloud.

Following the success of self-supervised Transformers in NLP and CV, Voxel-MAE utilizes a Transformer backbone for extracting point cloud features. 
The Transformer architecture is chosen as its pre-training scales favorably when deploying extensive masking, as only unmasked data are embedded in the encoder. 
Moreover, the model efficiently handles sparse point clouds by only processing non-empty voxels. 
Interestingly, only a handful of Transformer backbones exist for automotive point clouds \cite{Fan_2022_CVPR,mao2021voxel,pan20213d,pvt}, and their self-supervised pre-training has not been explored previously \cite{lu2022transformers}.
In this work, we use the Single-stride Sparse Transformer (SST) \cite{Fan_2022_CVPR} as our point cloud encoder, which applies a shifted-window transformer directly to the voxelized point cloud, similar to the Swin Transformer for images \cite{liu2021swin}. 
SST has achieved competitive results for 3D object detection, capturing fine details while being computationally efficient, making it a strong baseline to improve upon. 
For the pre-training, we follow the paradigm of MAE \cite{he2021masked} and equip the model with a lightweight decoder that is structurally similar to the encoder.

In summary, we present the following contributions:
\begin{itemize}
    \item We propose Voxel-MAE, a method for deploying MAE-style self-supervised pre-training on voxelized point clouds, and evaluate it on nuScenes, a large-scale automotive point cloud dataset. Our method is the first self-supervised pre-training scheme that uses a Transformer backbone for automotive point clouds.
    \item We tailor our method toward the voxel representation and use a unique set of reconstruction tasks to capture the characteristics of voxelized point clouds.
    \item We demonstrate that our method is data-efficient and reduces the need for annotated data. By pre-training, we outperform a fully-supervised equivalent when using only 40\% of the annotated data.
    \item Further, we show that Voxel-MAE boosts the performance of a Transformer-based detector by 1.75\%-points in mAP and 1.05\%-points in NDS, showcasing up to $2\times$ the performance increase compared to existing self-supervised methods.
\end{itemize}

\section{Related Work} 

\noindent \textbf{Masked autoencoders for language and images.} 
Masked language modeling (MLM) and its derivatives such as BERT \cite{devlin-etal-2019-bert} and GPT \cite{brown2020language,radford2018improving,radford2019language} have been very successful within NLP. These methods learn data representations by masking part of an input sentence and train models to predict the missing parts. The methods scale well, enabling training on datasets of unprecedented size and their representations generalize to various downstream tasks. Inspired by their success, multiple methods have applied similar techniques to the image domain \cite{bao2022beit,chen2020generative,dosovitskiy2021an,he2021masked,xie2022simmim}. Recently, the authors of \cite{he2021masked} proposed MAE, a simple approach where random image patches are masked and their pixel values are used as reconstruction targets. Further, they deploy an asymmetric encoder-decoder architecture, where only visible patches are embedded by the encoder, and a lightweight decoder is used for reconstruction. MAE is shown to improve performance on a range of downstream tasks compared to a fully supervised baseline. Voxel-MAE follows this design philosophy and makes the non-trivial translation to sparse point cloud data.

\begin{figure*}[th]
\begin{center}
\includegraphics[width=0.9\linewidth,trim={0cm 8.9cm 10.1cm 0cm},clip,page=2]{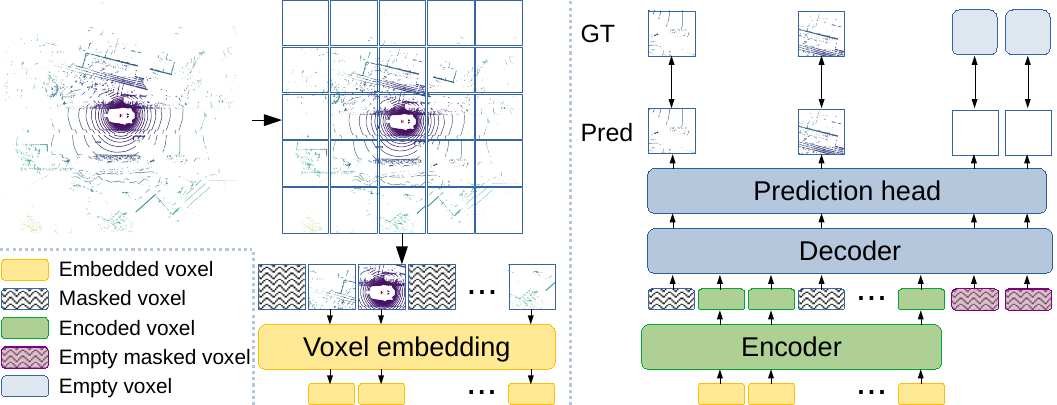}
\end{center}
   \caption{Our Voxel-MAE approach. First, the point cloud is voxelized with a fixed voxel size. The voxel size in the figure has been exaggerated for visualization purposes. During pre-training, a large subset (70\%) of the \textit{non-empty} voxels are masked out at random. The encoder is then applied only to the visible voxels, which are embedded using a dynamic voxel feature embedding \cite{zhou2020end}. Masked non-empty voxels and randomly selected empty voxels are embedded using the same learnable mask token. The sequence of mask tokens and encoded visible voxels are then processed by the decoder to reconstruct the masked point cloud and to discriminate between empty and non-empty voxels. After pre-training, the decoder is discarded and the encoder is applied to unmasked point clouds.}
\label{fig:voxel-mae}
\end{figure*}

\noindent \textbf{Masked autoencoders for point clouds.}
Inspired by the success of MLM in NLP and MAE in computer vision, multiple adaptations to the point cloud domain have been suggested. Point-BERT \cite{Yu_2022_CVPR} first introduced BERT-style pre-training for point clouds, masking and reconstructing parts of the input. While achieving competitive results, their approach relies on training a separate discrete Variational AutoEncoder (dVEA) for tokenizing point cloud patches, adding complexity and dependency on tokenizer performance. Point-MAE \cite{pang2022masked} removes the tokenizer and instead reconstructs the point patches directly, using the Chamfer distance for measuring the similarity between predicted and true point clouds. This speeds up training compared to Point-BERT and also improves downstream performance. MaskPoint \cite{liu2022masked} further speeds up pre-training by removing the point cloud reconstruction. Instead, the decoder is trained to discriminate between masked point patches and fake, empty ones, sampled at random.

\noindent \textbf{Self-supervised learning for 3D object detection.}
While outdoor 3D detection has much to gain from self-supervised learning, the field is generally under-explored. STRL \cite{huang2021spatio} follows the BYOL \cite{grill2020bootstrap} approach and trains two point cloud encoders to create consistent latent representations when presented with two temporally correlated point clouds. Training two encoders can however limit model size due to increased memory requirements during pre-training. GCC-3D \cite{liang2021exploring} applies contrastive learning by training models to produce voxel-wise similar features when presented with two augmented views of the same point cloud. In \cite{erccelik20223d}, the pre-training is done using two subsequent point clouds and the models are trained to estimate the scene flow between frames. This can be seen as a special case of masked autoencoder, where the masking is done temporally. However, their method relies on a special alternating training scheme, switching between self-supervised and supervised training. In contrast, our method enables a simpler, sequential training strategy where the models are first pre-trained and then fine-tuned as needed. Thus, we avoid issues where large unannotated datasets have to be processed each time the model is trained toward the downstream task.

\section{Methodology}

This work aims to extend the MAE-style pre-training \cite{he2021masked} to voxelized point clouds. The core idea remains to use an encoder to create rich latent representation from partial observations of the input, followed by a decoder to reconstruct the original input, as visualized in Fig.~\ref{fig:voxel-mae}. After pre-training, the encoder is used as a backbone for a 3D object detector. But, due to fundamental differences between images and point clouds, several modifications are needed for the effective training of Voxel-MAE, as outlined below.

\subsection{Masking and voxel embedding}
Similar to the division of images into non-overlapping patches, the point cloud is first divided into voxels. Voxels bring structure to the otherwise irregular point cloud, enabling efficient processing while retaining sufficient details for dense prediction tasks such as 3D OD. However, voxels also bring unique challenges compared to image patches.

First, a large fraction of the voxels in the field of view are generally empty due to occlusion and the inherent sparsity of lidar data. Rather than using all voxels, we discard empty voxels to avoid unnecessary computational strain. During pre-training, we mask a large fraction (70\%) of non-empty voxels and process only visible voxels with the encoder, further enhancing computational efficiency. The varying amount of visible voxels between scenes is handled elegantly by the many-to-many mapping of Transformers.

Second, due to the varying point density, the number of points assigned to individual voxels can vary from one to a few hundred. For embedding all points in each visible voxel to a single feature vector we use a dynamic voxel feature encoder \cite{zhou2020end}. Masked voxels are instead embedded with a shared, learnable mask token.

\subsection{Encoder}
For encoding the visible voxels we use the encoder of the Single-stride Sparse Transformer (SST) \cite{Fan_2022_CVPR}. SST is a Transformer-based 3D object detector operating on voxels, making it easy to transfer pre-trained backbone weights to the downstream task of 3D OD. The SST encoder is constructed by stacking multiple Transformer encoder layers, where non-empty voxels are treated as separate tokens and point clouds are considered to be sequences of such tokens. Further, each token is accompanied by a positional embedding based on the position of the voxel in the field of view. 

Since Transformers scale poorly with sequence length due to quadratic complexity in the self-attention mechanism, SST introduces regional grouping and regional shift. Inspired by the shifted windows in Swin Transformer \cite{liu2021swin}, the field of view is divided into non-overlapping 3D regions. Self-attention is only calculated among voxels within the same region, drastically reducing the computational load compared to global self-attention. To enable interaction between voxels from different regions, the regions are shifted every other encoder layer and voxels are grouped according to the new regions. The combination of regional grouping and only processing non-empty voxels limits the computational footprint of pre-training SST with Voxel-MAE, especially with extensive masking.

\subsection{Decoder}
After encoding the visible voxels, the decoder is used to leverage the rich latent representation for reconstructing the original point cloud. Note that the decoder is only used during pre-training and is discarded when fine-tuning the model toward downstream tasks. As can be seen in Fig.~\ref{fig:voxel-mae}, the sequence of embedded voxels is extended with the masked voxels. These are embedded as a shared, learned mask token along with their respective positional embedding, such that the decoder can distinguish between them. 

Besides the encoded and masked voxels, we also add a set of \textit{empty} masked voxels, similar to what is done in \cite{liu2022masked}. We do this by sampling randomly among the empty voxels in the field of view and embedding them in the same fashion as the non-empty, masked voxels. The empty masked voxels are added to make the reconstruction task harder and effectually promote the encoder's learning. By only processing voxels which contain points, the model would have close to perfect knowledge about occupancy, thus not having to learn about this property of point clouds. Instead, we force the decoder to learn to distinguish between non-empty and empty masked voxels and ignore empty voxels for reconstruction. Empirically, we found sampling 10\% of the empty voxels gave good performance, without introducing unnecessary computational overhead.

The decoder has a similar structure as the encoder, consisting of SST encoder layers, but using fewer layers. This can partially be motivated by the reduced time needed for pre-training, but we also find the encoder to achieve higher downstream task performance when trained in conjunction with a smaller decoder, similar to the results in \cite{he2021masked}.

\subsection{Reconstruction target} \label{sec:reconstruction}
The decoder is supervised with three different reconstruction tasks, each supervising a certain characteristic inherent to point clouds. For each task, we apply a separate linear layer to the decoder output to project the embedding to suitable dimensions. The three tasks and their corresponding loss functions are described below.

As mentioned previously, each voxel contains a varying number of points. For exact reconstruction, this would require the prediction heads to predict a different number of points for each voxel. This can be achieved using a Recurrent Neural Network for instance but at the cost of simplicity. Instead, we propose to predict a fixed number of points $n$, enabling the use of a simple linear layer for predicting said points. This reconstruction is supervised with the Chamfer distance, which measures the distance between two sets of points and allows the sets to have different cardinality. Let $P^{gt}=\{P_i^{gt}\}_{i=1}^N$ be the masked point cloud partitioned into $N$ voxels where each voxel $P_i^{gt}=\{\mathbf{x}_j\}_{j=1}^{n_i}$ contains $n_i$ points, where $n_i$ can vary between voxels. Similarly, the predicted point cloud $P^{pre}=\{P_i^{pre}\}_{i=1}^N$ contains $N$ voxels $P_i^{pre}=\{\hat{\mathbf{x}}_j\}_{j=1}^n$ with $n$ fixed for all $i$. We calculate the Chamfer distance for each masked voxel and define our Chamfer loss as 
\begin{equation}
    \begin{split}
    \mathcal{L}_c = \sum_{P_i^{pre} \in P^{pre}} \frac{1}{|P_i^{pre}|} \sum_{\hat{\mathbf{x}}\in P_i^{pre}} \min_{\mathbf{x}\in P_i^{gt}} ||\mathbf{x}-\hat{\mathbf{x}}||_2^2 +
    \\
    \sum_{P_i^{gt} \in P^{gt}} \frac{1}{|P_i^{gt}|} \sum_{{\mathbf{x}}\in P_i^{gt}} \min_{\hat{\mathbf{x}}\in P_i^{pre}} ||\mathbf{x}-\hat{\mathbf{x}}||_2^2.
    \end{split}
\end{equation}
When the number of predicted points $n$ exceeds the true number of points $n_i$ in a voxel, the model can still minimize the Chamfer loss by placing duplicate points in the same location. For the other scenario, $n<n_i$, it has been shown \cite{wu2021balanced} that the Chamfer loss encourages model predictions to capture details in the true point cloud even under cardinality mismatch.

For the model to further learn the uneven point cloud distribution explicitly we also predict the number of points $\hat{n}_i$ for each non-empty masked voxel. As the target $n_i$ can range from one to a few hundred, we supervise the prediction using the smooth L1 loss to avoid exploding gradients 
\begin{equation}
    \mathcal{L}_{np} = \begin{cases}
    \frac{(n_i-\hat{n}_i)^2}{2} & \text{if } |n_i-\hat{n}_i| < 1,
    \\
    |n_i-\hat{n}_i| - 0.5 & \text{otherwise.}
    \end{cases} 
\end{equation}

Lastly, for each masked voxel, we predict whether it is empty or non-empty. This task is supervised with a simple binary cross entropy loss $\mathcal{L}_{occ}$. The total loss for the pre-training is
\begin{equation}
    \label{eq:final_loss}
    \mathcal{L} = \alpha_{c}\mathcal{L}_{c} +
    \alpha_{np}\mathcal{L}_{np} +
    \alpha_{occ}\mathcal{L}_{occ},
\end{equation}
where $\alpha_{c},\alpha_{np},\alpha_{occ}$ are scalar weights for scaling each loss term. 

\section{Experiments} 

\begin{table*}[thb]
    \centering
    \begin{tabular}{c|c|cc|cccccccc}
        \toprule
        Dataset fraction & Pre-trained & mAP & NDS & ped. & car & truck & bus & barrier & T.C. & trailer & moto.\\
        \midrule
        \multirow{2}{*}{0.2} & \xmark & 42.43 & 55.60 & 73.5 & 78.6 & 42.5 & 49.5 & 55.1 & 38.9 & 18.9 & 41.6\\
        & \cmark & \textbf{47.35} & \textbf{59.06} & \textbf{78.4} & \textbf{80.8} & \textbf{47.7} & \textbf{58.9} & \textbf{60.5} & \textbf{46.1} & \textbf{22.2} & \textbf{45.1} \\
        \hdashline
        \multirow{2}{*}{0.4} & \xmark & 47.79 & 59.11 & 77.9 & \textbf{81.2} & 47.1 & 56.2 & 59.0 & 46.2 & 21.6 & \textbf{47.8}\\
        & \cmark & \textbf{50.02} & \textbf{61.01} & \textbf{81.3} & 80.3 & \textbf{49.6} & \textbf{61.1} & \textbf{62.6} & \textbf{49.4} & \textbf{24.1} & 47.7\\
        \hdashline
        \multirow{2}{*}{0.6} & \xmark & 47.77 & 59.57 & 78.0 & 81.2 & 46.3 & 59.1 & 58.0 & 46.5 & 24.2 & 49.4\\
         & \cmark & \textbf{51.00} & \textbf{61.76} & \textbf{81.4} & \textbf{81.9} & \textbf{50.6} & \textbf{60.8} & \textbf{63.1} & \textbf{52.9} & \textbf{25.1} & \textbf{53.1}\\
        \hdashline
        \multirow{2}{*}{0.8} & \xmark & 48.26 & 60.26 & 78.6 & 81.5 & 46.7 & 58.8 & 57.6 & 49.0 & 23.0 & 52.0\\
         & \cmark & \textbf{51.67} & \textbf{62.38} & \textbf{81.0} & \textbf{82.3} & \textbf{49.8} & \textbf{62.6} & \textbf{64.0} & \textbf{52.3} & \textbf{25.8} & \textbf{53.1}\\
        \hdashline
        \multirow{2}{*}{1.0} & \xmark & 49.08 & 60.75 & 78.5 & 81.9 & 47.8 & 60.1 & 59.7 & 48.7 & 23.1 & \textbf{51.6}\\
        & \cmark & \textbf{51.95} & \textbf{62.16} & \textbf{81.4} & \textbf{82.3} & \textbf{51.2} & \textbf{63.9} & \textbf{63.2} & \textbf{52.8} & \textbf{27.5} & 51.5\\
        \bottomrule
    \end{tabular}
    \caption{mAP, NDS, and AP per class on the nuScenes \textit{val} data for pre-trained and randomly initialized models when varying the amount of \textit{labeled} data. Pre-training and fine-tuning are done with \textit{ten} aggregated point cloud sweeps without intensity information. ped.=pedestrian. T.C.=traffic cone. moto.=motorcycle.}
    \label{tab:data_efficiency_10_sweeps}
\end{table*}

For our experiments, we use the popular self-driving dataset nuScenes  \cite{caesar2020nuscenes} which contains 1,000 sequences from Boston and Singapore, each sequence being 20 s long with raw data collected at 10 Hz, and annotations available at 2 Hz. Out of the 1,000 sequences, 850 are used for training and validation where we use established splits. All models are pre-trained on the raw training data. Following pre-training, models are fine-tuned toward the downstream task of 3D object detection.

SST \cite{Fan_2022_CVPR} was developed in the MMDetection3D framework \cite{mmdet3d2020} and originally evaluated in the Waymo Open Dataset \cite{sun2020scalability}. Due to inherent differences between the Waymo and nuScenes datasets, e.g., Waymo lidar having 64 lidar beams instead of 32, we extend the original SST implementation and tune hyperparameters for optimized nuScenes performance. For instance, we found that using a slightly larger voxel size and more encoder layers yields better performance than the original hyperparameters. Further, following standard practice, SST was trained using aggregated point cloud sweeps. For studying sensitivity to point cloud density, we evaluate our models with 2 and 10 sweeps, where results for 2 sweeps can be found in Section C in the supplementary material along with results from additional ablations. For complete training details see Section A in the supplementary material. 

\noindent\textbf{Pre-training.} Models are trained with the AdamW optimizer \cite{loshchilov2018decoupled} with $\beta_1=0.95$, $\beta_2=0.99$, and weight decay of 0.01. The initial learning rate is set to 5e-5 and gradually increased over the first 1000 iterations to 5e-4 and then decayed down to 1e-7 following a cosine annealing schedule. Pre-training is run on NVIDIA A100 for 200 epochs with a batch size of four. Loss weights are set as $\alpha_c=1$, $\alpha_{np}=0.1$, and $\alpha_{occ}=1$. The masking ratio is set to $0.7$ and non-empty voxels are sampled uniformly at random. Further, the point prediction head is set to predict 10 points. When calculating the Chamfer loss, we further limit the number of true points to fewer than 100 for computational efficiency, where points are selected at random. For remaining model details, see supplementary material Section B.

\noindent\textbf{Downstream task training.} When training toward the downstream task, weights for the voxel encoder and SST encoder layers are initialized either from pre-trained weights or randomly, depending on using Voxel-MAE or not. The remaining model parts are always initialized randomly. We use the AdamW optimizer with $\beta_1=0.9$, $\beta_2=0.999$, and weight decay of 0.05. The learning rate is increased from 1e-5 to 1e-3 during the first iterations and decreased with a cosine annealing schedule down to 1e-8. Models are trained for 288 epochs with a batch size of 4. 

\subsection{Data efficiency}
\begin{table*}[thb]
    \centering
    \begin{tabular}{c|c|cc|cccccccc}
        \toprule
        Labeled data & Unlabeled data & mAP & NDS & ped. & car & truck & bus & barrier & T.C. & trailer & moto.\\
        \midrule
        \multirow{6}{*}{0.01} & 0.0 & 8.03 & 19.37 & 15.4 & 45.1 & 4.6 & 1.0 & 11.3 & 2.8 & 0.0 & 0.0\\
         & 0.2 & 20.05 & 35.78 & 55.5 & 61.5 & 14.1 & 14.2 & 32.7 & 14.6 & 1.0 & 6.6\\
         & 0.4 & 20.75 & 34.75 & 58.7 & 62.4 & 14.5 & 13.8 & 35.6 & 15.2 & 0.8 & 6.4\\
         & 0.6 & 21.07 & 35.59 & 60.5 & 62.4 & 14.4 & 13.4 & 34.7 & 16.0 & 1.3 & 7.1\\
         & 0.8 & 21.23 & 36.74 & 60.3 & 63.9 & 15.4 & 13.1 & 36.2 & 14.7 & 1.0 & 7.4\\
         & 1.0 & 22.18 & 36.30 & 62.6 & 63.8 & 16.5 & 14.4 & 38.3 & 17.6 & 0.9 & 6.9\\
        \hdashline
        \multirow{6}{*}{0.05} & 0.0 & 24.99 & 42.77 & 56.6 & 68.5 & 22.4 & 18.2 & 36.6 & 22.0 & 5.5 & 16.6 \\
         & 0.2 & 33.93 & 49.58 & 71.2 & 74.1 & 33.4 & 34.6 & 48.1 & 34.1 & 11.9 & 21.7 \\
         & 0.4 & 34.88 & 49.87 & 72.1 & 74.7 & 33.5 & 37.1 & 49.2 & 34.6 & 13.2 & 23.9 \\
         & 0.6 & 34.74 & 49.95 & 72.7 & 74.5 & 34.1 & 36.5 & 47.7 & 36.7 & 11.1 & 22.4\\
         & 0.8 & 35.07 & 49.98 & 72.4 & 74.6 & 35.0 & 34.7 & 49.2 & 35.7 & 10.6 & 25.1\\
         & 1.0 & 36.01 & 50.91 & 73.8 & 74.9 & 34.1 & 38.2 & 50.8 & 37.6 & 11.2 & 23.9\\
        \bottomrule
    \end{tabular}
    \caption{mAP, NDS, and AP per class on the nuScenes \textit{val} data for pre-trained and randomly initialized models when varying the amount of \textit{unlabeled} data. 0.0 refers to the model trained from scratch. Pre-training and fine-tuning are done with ten aggregated point cloud sweeps without intensity information. ped.=pedestrian. T.C.=traffic cone. moto.=motorcycle.}
    \label{tab:data_efficiency_10_sweeps_varying_unlabeled}
\end{table*}

\noindent\textbf{Varying amount of labeled data.} One of the major benefits of using self-supervised learning is a reduced need for annotated data. To study the effects of various dataset sizes we train SST with and without Voxel-MAE with varying fractions of the annotated dataset held out. Specifically, we use $\{0.2,0.4,0.6,0.8,1.0\}$ of the annotated dataset for training the 3D OD models, where one model is initialized randomly and one has been pre-trained on the Voxel-MAE tasks. Pre-training was done on the entire nuScenes training dataset. To determine which scenes to use in each fraction, the training dataset was sorted based on scene timestamps. Then, scenes were chosen based on their index modulus 5, e.g., for extracting 20\% of the dataset, all scenes with index $i$ were chosen if $i \mod 5 = 0$, while for 40\% we used $i \mod 5 \in \{0,2\}$ as our selection criteria. This way, the temporal dependency between frames is minimized and the reduced datasets have similar diversity as the entire dataset. We report mAP and NDS scores for the nuScenes validation set in Table \ref{tab:data_efficiency_10_sweeps}.

From Table \ref{tab:data_efficiency_10_sweeps} we can see that by training SST from scratch with randomly initialized weights, the model achieves 49.08 mAP and 60.75 NDS when using 10 aggregated point cloud sweeps. In comparison, the pre-trained model, using only 40\% of the annotated data, achieves 50.02 mAP and 61.01 NDS, hence outperforming the version without pre-training. The substantial gap of 1 mAP point indicates that even less than 40\% of the annotated data would suffice.

The largest performance increase for Voxel-MAE in comparison to the baseline can be seen when fine-tuning on the smallest fraction of annotated data. In those instances, mAP is increased by close to 5 mAP points and NDS by $\sim 3.5$ points. The low-data regime benefits the most from expressive pre-trained voxel features. Nonetheless, pre-training with Voxel-MAE consistently outperforms the randomly initialized equivalent, even as the entire annotated dataset is used. Naturally, the performance gap shrinks as more annotated data is used, but the gap remains large regardless of the fraction of annotated data. For instance, using all of the annotated data, Voxel-MAE results in a 2.87 mAP point and a 1.41 NDS point increase. This indicates that our pre-training is useful for learning general point cloud representations which improve both data efficiency and final performance for existing methods.

\noindent\textbf{Varying amount of unlabeled data.} We also study the effect of varying the amount of unlabeled data when keeping the amount of labeled data fixed. This simulates the scenario where the amount of unlabeled data is much greater than the amount of labeled data, which is generally the case for real-world applications. For this, we pre-train five models on varying fractions of the entire dataset, namely $\{0.2,0.4,0.6,0.8,1.0\}$, where 1.0 is equivalent to using all available data. Next, models are fine-tuned on 1\% and 5\% of the annotated data. Their results, compared to a model without pre-training, are shown in Table \ref{tab:data_efficiency_10_sweeps_varying_unlabeled}.

All models pre-trained with Voxel-MAE outperform their corresponding baseline. Note that already using 20\% of the data for pre-training brings large increases in mAP and NDS, e.g., a 12 mAP point and 16.4 NDS point increment when using 1\% of the labeled data for fine-tuning. Further, performance increases as the amount of unlabeled data grow, showing that our proposed method makes effective use of large unannotated datasets. We also note that the increment in performance between some levels of unlabeled data might seem minor, compared to the step from no-pretraining to using 20\% of the data. We believe this to be an effect of the nuScenes dataset and our selection method when holding out part of the data. For the 0.2 dataset, we select frames uniformly in time, minimizing their temporal correlation. For the larger fractions, we only add frames that are already close in time to the ones contained in the 0.2 dataset, as nuScenes consists of sequence data. This limits diversity and the amount of new information being added when increasing the size of the pre-training dataset.

\subsection{Comparison to SOTA self-supervised learning methods}

While self-supervised pre-training recently has enjoyed much attention for point clouds in general, only a handful of methods have been proposed for improving automotive 3D OD performance. For nuScenes, two self-supervised techniques have been evaluated prior to this work. In \cite{liang2021exploring}, a voxel-based CNN backbone is trained to create consistent latent features for two different views of the same scene using contrastive learning.  In \cite{erccelik20223d}, the model is instead supervised to estimate the scene flow, i.e., the location of points in the consecutive frame. Further, \cite{erccelik20223d} deploys a custom training scheme, where the training objective is altered between the self-supervised task and the object detection task. 

We compare Voxel-MAE to existing methods in Table \ref{tab:selfsupervised_methods}. Note that the models use different types of backbones, which can affect the comparison. We report mAP and NDS both for models trained from scratch and the ones pre-trained with the various self-supervised techniques. For SST, we use 10 aggregated sweeps. Further, we trained a separate version that includes intensity information for each point in both pre-training and fine-tuning, something that was omitted in the original implementation \cite{Fan_2022_CVPR}. We found this to help final detection performance compared to the intensity-free version, while Voxel-MAE still shows a substantial increase compared to the baseline.

\begin{table}[thb]
    \centering
    \begin{tabular}{cll}
        \toprule
        Method & mAP & NDS \\
        \midrule
        PointPillars  \cite{erccelik20223d} & 40.02 & 53.29  \\
        $[$S$]$ PointPillars \cite{erccelik20223d} & $42.06^{+2.04}$ & $55.02^{\mathbf{+1.73}}$ \\ \hdashline
        CenterPoint + PP \cite{erccelik20223d} & 49.13 & 59.73 \\
        $[$S$]$ CenterPoint + PP  \cite{erccelik20223d} & $49.89^{+0.76}$ & $60.01^{+0.28}$ \\
        \hdashline
        CenterPoint + PP \cite{liang2021exploring} & 49.61 & 60.20 \\ 
        $[$S$]$ CenterPoint + PP \cite{liang2021exploring} & $50.84^{+1.23}$ & $60.76^{+0.56}$ \\\hdashline
        CenterPoint + V \cite{liang2021exploring} & 56.19 & 64.48 \\
        $[$S$]$ CenterPoint + V \cite{liang2021exploring} & $57.26^{+1.07}$ & $65.01^{+0.53}$ \\
        \hdashline
        SST & 49.08 & 60.75 \\
        $[$S$]$ SST + Voxel-MAE & $51.95^{\mathbf{+2.87}}$ & $62.16^{{+1.41}}$ \\ \hdashline
        SST* & 53.39 & 62.95 \\
        $[$S$]$ SST* + Voxel-MAE & $55.14^{+1.75}$ & $64.00^{+1.05}$ \\
        \bottomrule
    \end{tabular}
    \caption{Detection performance on the nuScenes \textit{val} data for SOTA self-supervised methods. $[$S$]$ indicates models have been pre-trained with a self-supervised task, while other models have been randomly initialized. PP = PointPillars. V = VoxelNet. SST* uses intensity information.}
    \label{tab:selfsupervised_methods}
\end{table}

In Table \ref{tab:selfsupervised_methods}, we see the largest increase in mAP when pre-training SST with Voxel-MAE. 
Performance in terms of NDS is increased the most for the PointPillars model trained by \cite{erccelik20223d}. 
This is, however, also the worst-performing detector, i.e., the baseline from which it is the easiest to improve upon. 
By instead comparing SST to the similar performing CenterPoint models trained by \cite{liang2021exploring} and \cite{erccelik20223d}, we see the effectiveness of our proposed Voxel-MAE approach. 
For the SST model with intensity information, our absolute performance gains (measured in NDS) are almost twice those for the best-performing existing methods. 
This highlights the potential for self-supervised Transformers in the point cloud domain, results that are in line with what has been observed in fields such as NLP and computer vision. 

\subsection{Comparison to SOTA object detectors}
Transformers are widely used for NLP and computer vision tasks, and for automotive data, it has also been used for camera-lidar fusion \cite{bai2022transfusion,zeng2022lift} or for camera-only 3D OD \cite{li2022bevformer,liu2022petrv2}. 
Nonetheless, Transformers remain relatively under-explored as a backbone for lidar data. 
Pointformer \cite{pan20213d} is the only Transformer-based lidar object detector that has been used on the nuScenes dataset and we compare its performance to SST in Table \ref{tab:transformer_backbones}. 
We can see that pre-training with Voxel-MAE, SST can outperform Pointformer by a substantial margin. 
However, Transformer backbones are still lagging behind CNN-based feature extractors. 
We hope that our work can encourage further research toward the use of Transformer backbones for automotive point clouds, to make use of vast amounts of raw data.

\begin{table}[thb]
    \centering
    \begin{tabular}{ccc}
        \toprule
        Backbone & Method & mAP \\
        \midrule
        \multirow{4}{*}{CNN} & CenterPoint \cite{liang2021exploring} & 56.19 \\
        & UVTR-L \cite{li2022uvtr} &  60.90 \\
        & VISTA \cite{deng2022vista} & 62.83 \\
        & TransFusion-L \cite{bai2022transfusion} & 65.19 \\
        \hdashline
        \multirow{3}{*}{Transformer} & Pointformer \cite{pan20213d} &  53.60 \\ 
        & SST*  & 53.39 \\
        & SST* + Voxel-MAE & 55.14 \\ \bottomrule
    \end{tabular}
    \caption{Detection performance on the nuScenes \textit{val} dataset, comparing SST to SOTA 3D OD methods. SST* uses intensity information.}
    \label{tab:transformer_backbones}
\end{table}

\subsection{Loss ablation}
One of the key differences between Voxel-MAE and MAE for images \cite{he2021masked}, is the reconstruction task and its accompanying losses. 
These differences stem from the inherent sparsity of point cloud data, which is not present in regular images. 
We study the effects of our proposed losses in Table \ref{tab:loss_ablation}. 
From the table, we see that a naive extension of MAE, i.e., only predicting points and supervising with the Chamfer loss, actually hurts mAP performance compared to no pre-training. 
Further, the table suggests the best loss combination is using Chamfer and empty voxels, or the number of points and empty voxels, depending on if mAP or NDS is selected as the criterion. 
From these results, we believe that by tuning $\alpha_c$, $\alpha_{np}$ and $\alpha_{occ}$ in \eqref{eq:final_loss} one can likely achieve the highest performance in both mAP and NDS when using all three losses. 
Tuning for maximum performance was however considered out of scope in this work.

\begin{table}[htb]
    \centering
    \begin{tabular}{ccc|cc}
        \toprule
        Chamfer & \#Points & Empty & mAP & NDS  \\
        \midrule
        \rowcolor{Gainsboro!60} & & & 43.60 & 55.19 \\ 
        \cmark & & & 43.46 & 55.29 \\ 
        \rowcolor{Gainsboro!60} & \cmark  & & 44.27 & 55.83 \\ 
        \cmark & \cmark  & & 45.03 & 56.10 \\ 
        \rowcolor{Gainsboro!60}\cmark & & \cmark  & 44.82 & \textbf{56.25} \\
        & \cmark & \cmark & \textbf{45.25} & 56.11 \\ 
        \rowcolor{Gainsboro!60}\cmark & \cmark & \cmark & 44.62 & 56.00 \\
        \bottomrule
    \end{tabular}
    \caption{mAP and NDS on the nuScenes \textit{val} data for pre-trained models for different combinations of reconstruction tasks. Pre-training and fine-tuning are done with \textit{two} aggregated point cloud sweeps without intensity information.}
    \label{tab:loss_ablation}
\end{table}

\subsection{Qualitative evaluation}
\begin{figure}[th]
    \centering
    \includegraphics[width=0.49\linewidth,trim={31.8cm 27.9cm 25.5cm 30.6cm},clip,page=1]{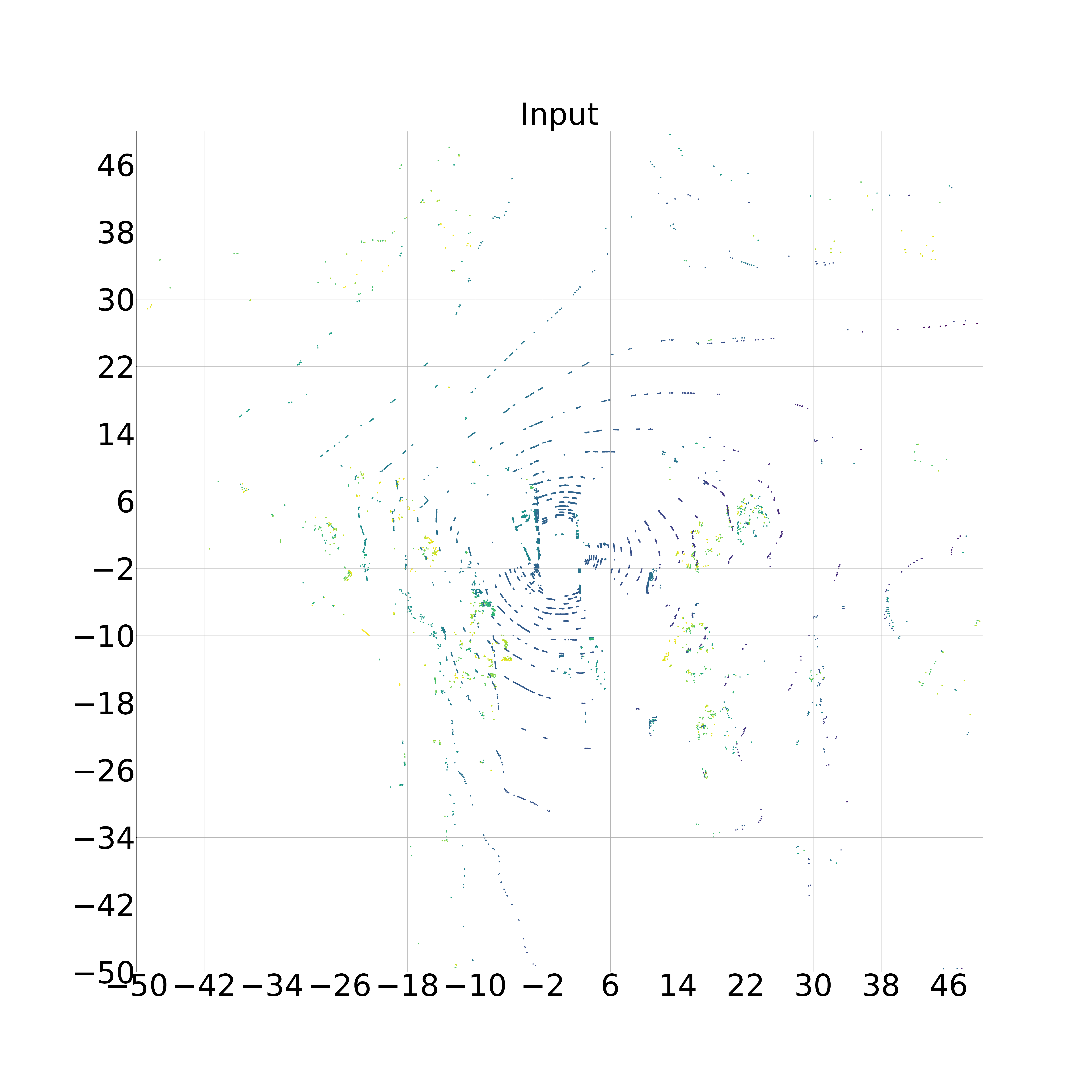}
    \includegraphics[width=0.49\linewidth,trim={31.8cm 27.9cm 25.5cm 30.6cm},clip,page=1]{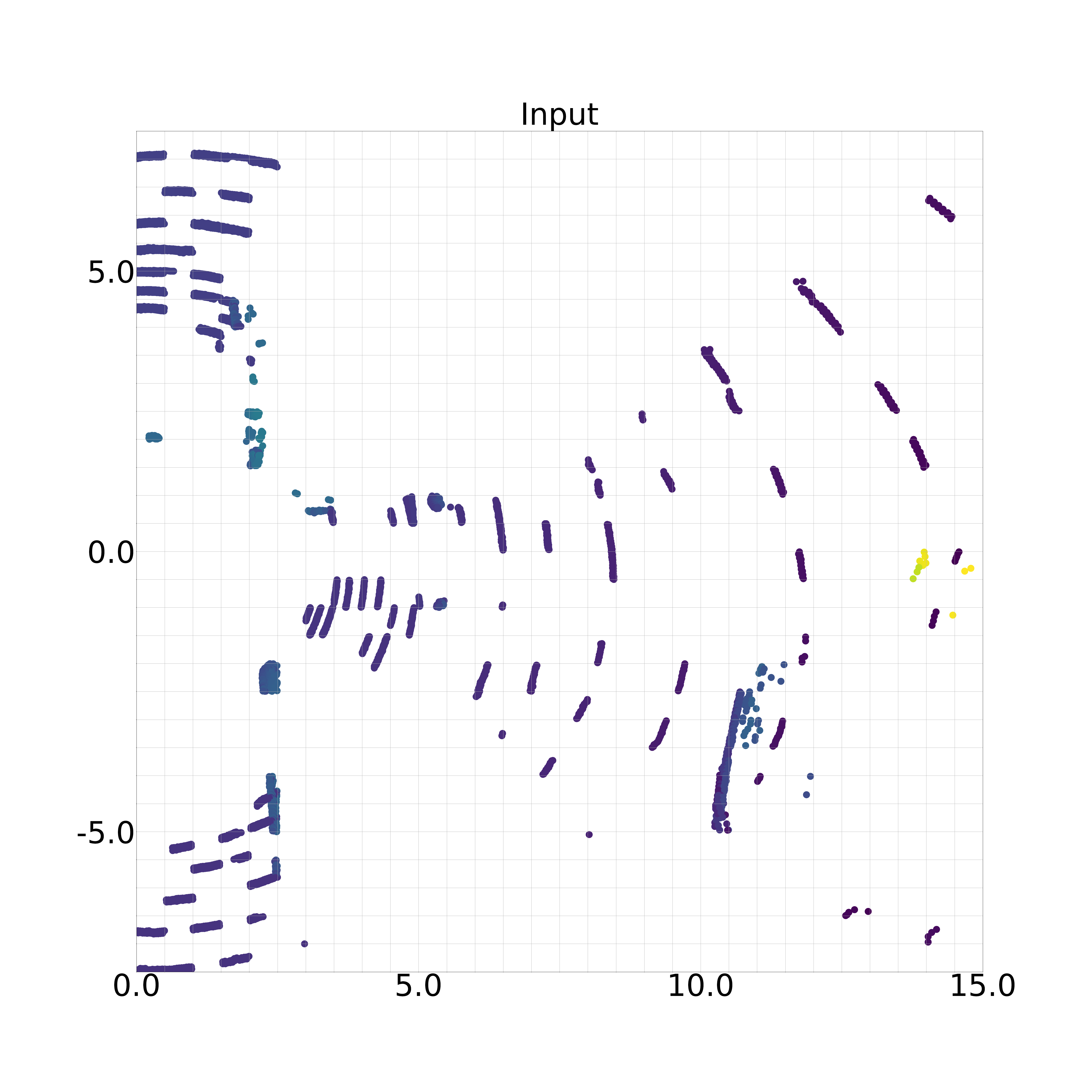}
    \includegraphics[width=0.49\linewidth,trim={31.8cm 27.9cm 25.5cm 30.6cm},clip,page=1]{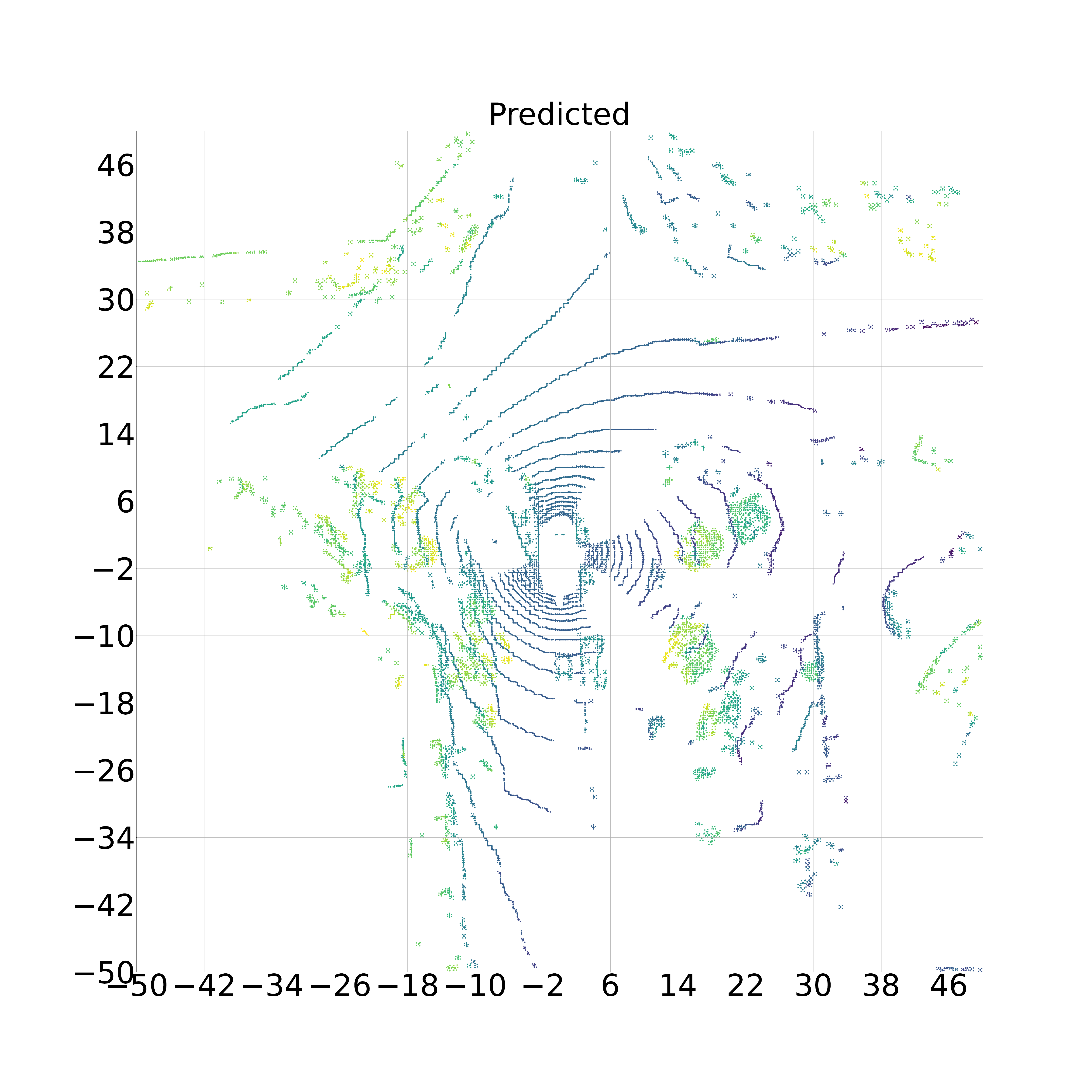}
    \includegraphics[width=0.49\linewidth,trim={31.8cm 27.9cm 25.5cm 30.6cm},clip,page=1]{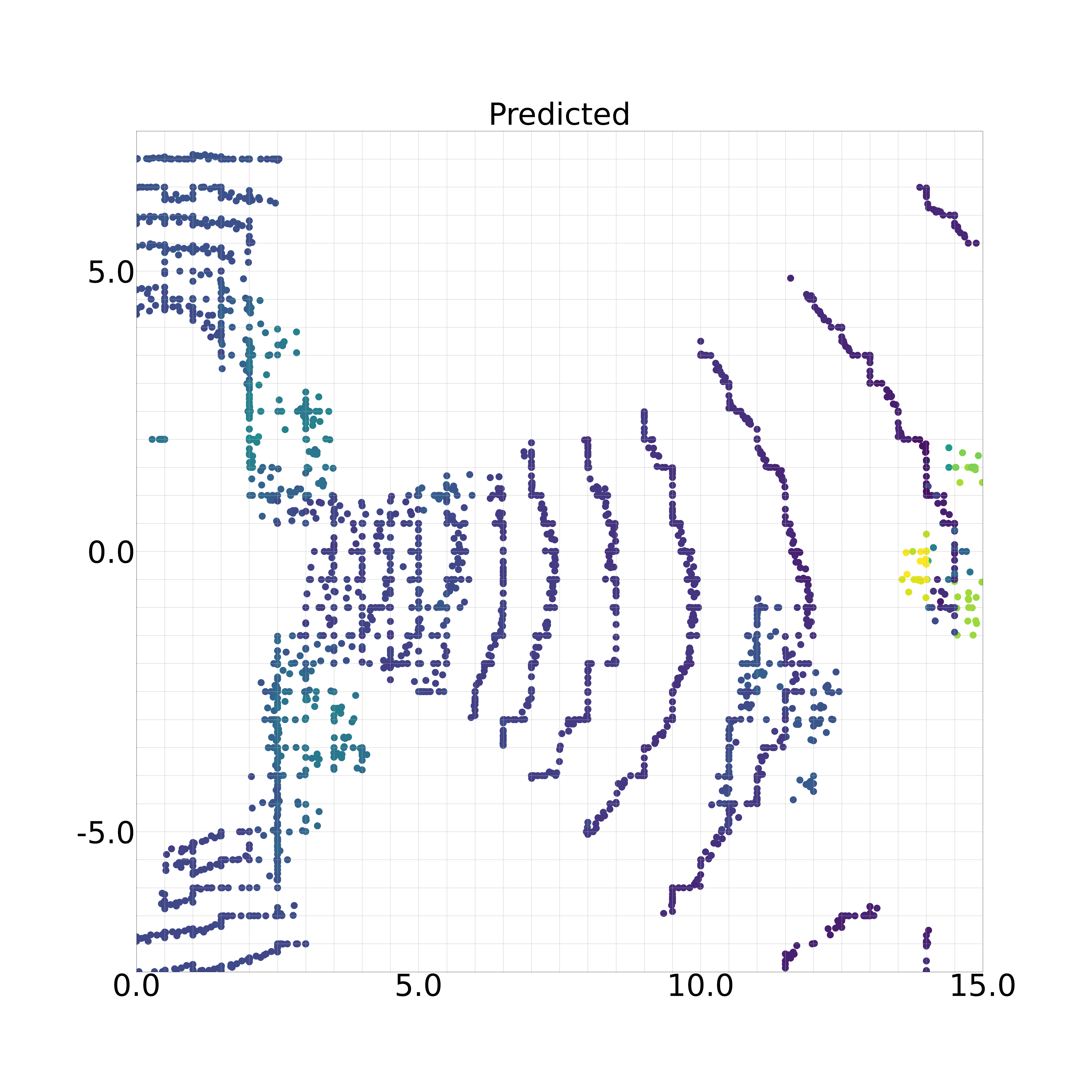}
    \includegraphics[width=0.49\linewidth,trim={31.8cm 27.9cm 25.5cm 30.6cm},clip,page=1]{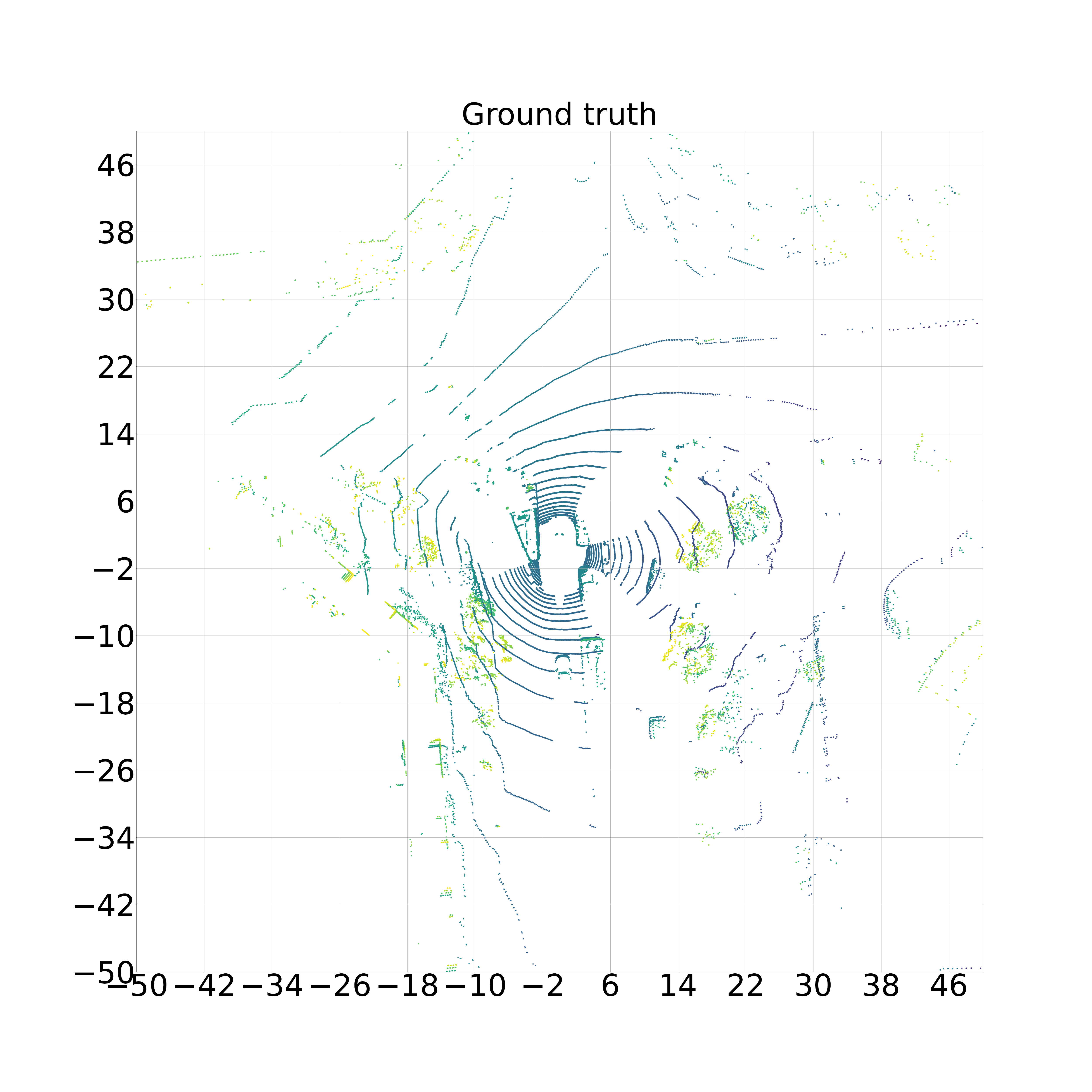}
    \includegraphics[width=0.49\linewidth,trim={31.8cm 27.9cm 25.5cm 30.6cm},clip,page=1]{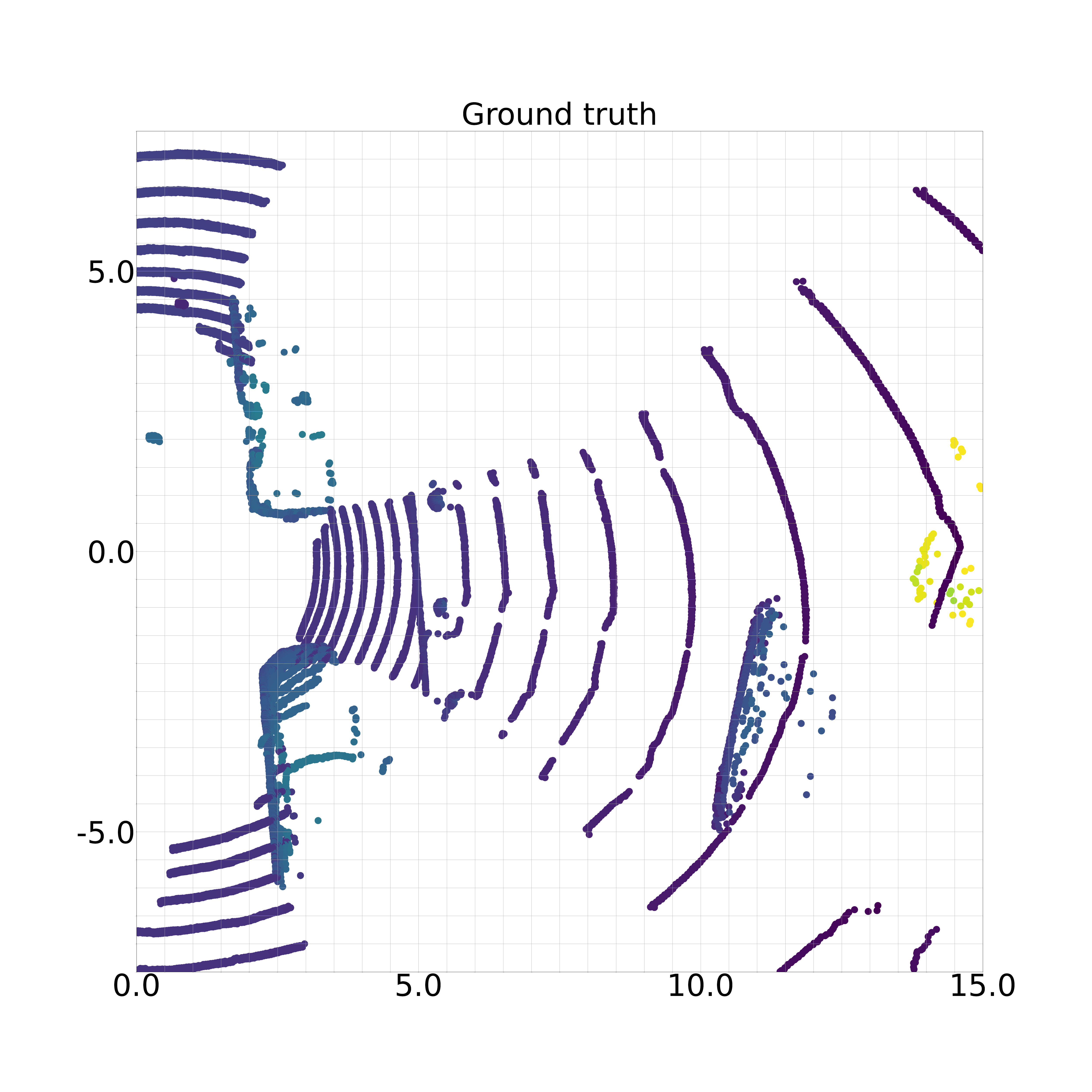}
    \caption{Masked (top), reconstructed (middle) and true point cloud (bottom). Left shows the entire field of view ($50\times50$ meters). Right shows a zoomed-in version ($x\in[0,15],y\in[-7.5,7.5]$) of the same scene, where the grid represents the voxels. The masking ratio is 70\% of non-empty voxels. Color represents points' height, purple being ground level and yellow being maximum height. We include reconstructed unmasked voxels for visualization purposes, although the model has not been supervised for this.}
    \label{fig:reconstruction}
\end{figure}

Figure \ref{fig:reconstruction} shows examples of a reconstructed point cloud from the nuScenes validation set. The model displays an understanding of general shapes, predicts reasonable height for most points, and captures the characteristic lidar lines along the ground plane. Note that the model does this for a \textit{single} sweep point cloud, which it has not been trained on.

\section{Conclusions} 
We propose Voxel-MAE, a simple masked point modeling pre-training paradigm tailored toward voxelized point clouds. Experiments on a large-scale automotive dataset show that Voxel-MAE learns useful point cloud representations from raw lidar point clouds. Our method yields a notable performance increase for a competitive Transformer-based 3D object detector. Further, our pre-training reduces the need for annotated data, enabling us to achieve competitive detection performance when using a fraction of available annotations. We hope our work can encourage further research on Transformers for automotive data. 

\textbf{Future directions} include studies of temporal masking, similar to methods in the video domain \cite{feichtenhofer2022masked,tong2022videomae}, to learn both spatial and temporal representations useful for multi-object tracking and motion prediction. 

\textbf{Broader impact.}
Self-supervised learning in general, and our method in particular, enable the utilization of otherwise unused data, opening up for energy-consuming training of ever-larger models and potentially requiring the storage of huge datasets. Associated resources can have a negative environmental impact and also limit the development and deployment of these models to well-funded actors.  

\section*{Acknowledgements}
This work was partially supported by the Wallenberg AI, Autonomous Systems and Software Program (WASP) funded by the Knut and Alice Wallenberg Foundation. The experiments were enabled by resources provided by the Swedish National Infrastructure for Computing (SNIC) at Chalmers Centre for Computational Science and Engineering (C3SE) and National Supercomputer Centre (NSC) at Linköping University, partially funded by the Swedish Research Council through grant agreement no. 2018-05973.

\FloatBarrier

{\small
\bibliographystyle{ieee_fullname}
\bibliography{bib}

\begin{thebibliography}{10}\itemsep=-1pt

\bibitem{bai2022transfusion}
Xuyang Bai, Zeyu Hu, Xinge Zhu, Qingqiu Huang, Yilun Chen, Hongbo Fu, and
  Chiew-Lan Tai.
\newblock Transfusion: Robust lidar-camera fusion for 3d object detection with
  transformers.
\newblock In {\em Proceedings of the IEEE/CVF Conference on Computer Vision and
  Pattern Recognition}, pages 1090--1099, 2022.

\bibitem{bao2022beit}
Hangbo Bao, Li Dong, Songhao Piao, and Furu Wei.
\newblock {BE}it: {BERT} pre-training of image transformers.
\newblock In {\em International Conference on Learning Representations}, 2022.

\bibitem{brown2020language}
Tom Brown, Benjamin Mann, Nick Ryder, Melanie Subbiah, Jared~D Kaplan, Prafulla
  Dhariwal, Arvind Neelakantan, Pranav Shyam, Girish Sastry, Amanda Askell,
  et~al.
\newblock Language models are few-shot learners.
\newblock {\em Advances in neural information processing systems},
  33:1877--1901, 2020.

\bibitem{caesar2020nuscenes}
Holger Caesar, Varun Bankiti, Alex~H Lang, Sourabh Vora, Venice~Erin Liong,
  Qiang Xu, Anush Krishnan, Yu Pan, Giancarlo Baldan, and Oscar Beijbom.
\newblock nuscenes: A multimodal dataset for autonomous driving.
\newblock In {\em Proceedings of the IEEE/CVF conference on computer vision and
  pattern recognition}, pages 11621--11631, 2020.

\bibitem{caron2021emerging}
Mathilde Caron, Hugo Touvron, Ishan Misra, Herv{\'e} J{\'e}gou, Julien Mairal,
  Piotr Bojanowski, and Armand Joulin.
\newblock Emerging properties in self-supervised vision transformers.
\newblock In {\em Proceedings of the IEEE/CVF International Conference on
  Computer Vision}, pages 9650--9660, 2021.

\bibitem{shapenet2015}
Angel~X. Chang, Thomas Funkhouser, Leonidas Guibas, Pat Hanrahan, Qixing Huang,
  Zimo Li, Silvio Savarese, Manolis Savva, Shuran Song, Hao Su, Jianxiong Xiao,
  Li Yi, and Fisher Yu.
\newblock {ShapeNet: An Information-Rich 3D Model Repository}.
\newblock Technical Report arXiv:1512.03012 [cs.GR], Stanford University ---
  Princeton University --- Toyota Technological Institute at Chicago, 2015.

\bibitem{chen2020generative}
Mark Chen, Alec Radford, Rewon Child, Jeffrey Wu, Heewoo Jun, David Luan, and
  Ilya Sutskever.
\newblock Generative pretraining from pixels.
\newblock In {\em International Conference on Machine Learning}, pages
  1691--1703. PMLR, 2020.

\bibitem{chen2020simple}
Ting Chen, Simon Kornblith, Mohammad Norouzi, and Geoffrey Hinton.
\newblock A simple framework for contrastive learning of visual
  representations.
\newblock In {\em International conference on machine learning}, pages
  1597--1607. PMLR, 2020.

\bibitem{mmdet3d2020}
MMDetection3D Contributors.
\newblock {MMDetection3D: OpenMMLab} next-generation platform for general {3D}
  object detection.
\newblock \url{https://github.com/open-mmlab/mmdetection3d}, 2020.

\bibitem{dai2017scannet}
Angela Dai, Angel~X Chang, Manolis Savva, Maciej Halber, Thomas Funkhouser, and
  Matthias Nie{\ss}ner.
\newblock Scannet: Richly-annotated 3d reconstructions of indoor scenes.
\newblock In {\em Proceedings of the IEEE conference on computer vision and
  pattern recognition}, pages 5828--5839, 2017.

\bibitem{deng2022vista}
Shengheng Deng, Zhihao Liang, Lin Sun, and Kui Jia.
\newblock Vista: Boosting 3d object detection via dual cross-view spatial
  attention.
\newblock In {\em Proceedings of the IEEE Conference on Computer Vision and
  Pattern Recognition}, 2022.

\bibitem{devlin-etal-2019-bert}
Jacob Devlin, Ming-Wei Chang, Kenton Lee, and Kristina Toutanova.
\newblock {BERT}: Pre-training of deep bidirectional transformers for language
  understanding.
\newblock In {\em Proceedings of the 2019 Conference of the North {A}merican
  Chapter of the Association for Computational Linguistics: Human Language
  Technologies, Volume 1 (Long and Short Papers)}, pages 4171--4186,
  Minneapolis, Minnesota, June 2019. Association for Computational Linguistics.

\bibitem{dosovitskiy2021an}
Alexey Dosovitskiy, Lucas Beyer, Alexander Kolesnikov, Dirk Weissenborn,
  Xiaohua Zhai, Thomas Unterthiner, Mostafa Dehghani, Matthias Minderer, Georg
  Heigold, Sylvain Gelly, Jakob Uszkoreit, and Neil Houlsby.
\newblock An image is worth 16x16 words: Transformers for image recognition at
  scale.
\newblock In {\em International Conference on Learning Representations}, 2021.

\bibitem{erccelik20223d}
Eme{\c{c}} Er{\c{c}}elik, Ekim Yurtsever, Mingyu Liu, Zhijie Yang, Hanzhen
  Zhang, P{\i}nar Top{\c{c}}am, Maximilian Listl, Y{\i}lmaz~Kaan
  {\c{C}}ayl{\i}, and Alois Knoll.
\newblock 3d object detection with a self-supervised lidar scene flow backbone.
\newblock In {\em Proceedings of the European Conference on Computer Vision
  (ECCV)}, 2022.

\bibitem{Fan_2022_CVPR}
Lue Fan, Ziqi Pang, Tianyuan Zhang, Yu-Xiong Wang, Hang Zhao, Feng Wang, Naiyan
  Wang, and Zhaoxiang Zhang.
\newblock Embracing single stride 3d object detector with sparse transformer.
\newblock In {\em Proceedings of the IEEE/CVF Conference on Computer Vision and
  Pattern Recognition (CVPR)}, pages 8458--8468, June 2022.

\bibitem{feichtenhofer2022masked}
Christoph Feichtenhofer, Haoqi Fan, Yanghao Li, and Kaiming He.
\newblock Masked autoencoders as spatiotemporal learners.
\newblock {\em arXiv preprint arXiv:2205.09113}, 2022.

\bibitem{fu2022pos}
Kexue Fu, Peng Gao, ShaoLei Liu, Renrui Zhang, Yu Qiao, and Manning Wang.
\newblock Pos-bert: Point cloud one-stage bert pre-training.
\newblock {\em arXiv preprint arXiv:2204.00989}, 2022.

\bibitem{grill2020bootstrap}
Jean-Bastien Grill, Florian Strub, Florent Altch{\'e}, Corentin Tallec, Pierre
  Richemond, Elena Buchatskaya, Carl Doersch, Bernardo Avila~Pires, Zhaohan
  Guo, Mohammad Gheshlaghi~Azar, et~al.
\newblock Bootstrap your own latent-a new approach to self-supervised learning.
\newblock {\em Advances in Neural Information Processing Systems},
  33:21271--21284, 2020.

\bibitem{he2021masked}
Kaiming He, Xinlei Chen, Saining Xie, Yanghao Li, Piotr Doll\'ar, and Ross
  Girshick.
\newblock Masked autoencoders are scalable vision learners.
\newblock In {\em Proceedings of the IEEE/CVF Conference on Computer Vision and
  Pattern Recognition (CVPR)}, pages 16000--16009, June 2022.

\bibitem{huang2021spatio}
Siyuan Huang, Yichen Xie, Song-Chun Zhu, and Yixin Zhu.
\newblock Spatio-temporal self-supervised representation learning for 3d point
  clouds.
\newblock In {\em Proceedings of the IEEE/CVF International Conference on
  Computer Vision}, pages 6535--6545, 2021.

\bibitem{li2022uvtr}
Yanwei Li, Yilun Chen, Xiaojuan Qi, Zeming Li, Jian Sun, and Jiaya Jia.
\newblock Unifying voxel-based representation with transformer for 3d object
  detection.
\newblock In {\em Advances in Neural Information Processing Systems}, 2022.

\bibitem{li2022bevformer}
Zhiqi Li, Wenhai Wang, Hongyang Li, Enze Xie, Chonghao Sima, Tong Lu, Qiao Yu,
  and Jifeng Dai.
\newblock Bevformer: Learning bird's-eye-view representation from multi-camera
  images via spatiotemporal transformers.
\newblock In {\em Proceedings of the European Conference on Computer Vision
  (ECCV)}, 2022.

\bibitem{liang2021exploring}
Hanxue Liang, Chenhan Jiang, Dapeng Feng, Xin Chen, Hang Xu, Xiaodan Liang, Wei
  Zhang, Zhenguo Li, and Luc Van~Gool.
\newblock Exploring geometry-aware contrast and clustering harmonization for
  self-supervised 3d object detection.
\newblock In {\em Proceedings of the IEEE/CVF International Conference on
  Computer Vision}, pages 3293--3302, 2021.

\bibitem{liu2022masked}
Haotian Liu, Mu Cai, and Yong~Jae Lee.
\newblock Masked discrimination for self-supervised learning on point clouds.
\newblock In {\em Proceedings of the European Conference on Computer Vision
  (ECCV)}, 2022.

\bibitem{liu2022petrv2}
Yingfei Liu, Junjie Yan, Fan Jia, Shuailin Li, Qi Gao, Tiancai Wang, Xiangyu
  Zhang, and Jian Sun.
\newblock Petrv2: A unified framework for 3d perception from multi-camera
  images.
\newblock {\em arXiv preprint arXiv:2206.01256}, 2022.

\bibitem{liu2021swin}
Ze Liu, Yutong Lin, Yue Cao, Han Hu, Yixuan Wei, Zheng Zhang, Stephen Lin, and
  Baining Guo.
\newblock Swin transformer: Hierarchical vision transformer using shifted
  windows.
\newblock In {\em Proceedings of the IEEE/CVF International Conference on
  Computer Vision}, pages 10012--10022, 2021.

\bibitem{loshchilov2018decoupled}
Ilya Loshchilov and Frank Hutter.
\newblock Decoupled weight decay regularization.
\newblock In {\em International Conference on Learning Representations}, 2019.

\bibitem{lu2022transformers}
Dening Lu, Qian Xie, Mingqiang Wei, Linlin Xu, and Jonathan Li.
\newblock Transformers in 3d point clouds: A survey.
\newblock {\em arXiv preprint arXiv:2205.07417}, 2022.

\bibitem{mao2021voxel}
Jiageng Mao, Yujing Xue, Minzhe Niu, Haoyue Bai, Jiashi Feng, Xiaodan Liang,
  Hang Xu, and Chunjing Xu.
\newblock Voxel transformer for 3d object detection.
\newblock In {\em Proceedings of the IEEE/CVF International Conference on
  Computer Vision}, pages 3164--3173, 2021.

\bibitem{pan20213d}
Xuran Pan, Zhuofan Xia, Shiji Song, Li~Erran Li, and Gao Huang.
\newblock 3d object detection with pointformer.
\newblock In {\em Proceedings of the IEEE/CVF Conference on Computer Vision and
  Pattern Recognition}, pages 7463--7472, 2021.

\bibitem{pang2022masked}
Yatian Pang, Wenxiao Wang, Francis~EH Tay, Wei Liu, Yonghong Tian, and Li Yuan.
\newblock Masked autoencoders for point cloud self-supervised learning.
\newblock In {\em Proceedings of the European Conference on Computer Vision
  (ECCV)}, 2022.

\bibitem{radford2018improving}
Alec Radford, Karthik Narasimhan, Tim Salimans, and Ilya Sutskever.
\newblock Improving language understanding by generative pre-training.
\newblock 2018.

\bibitem{radford2019language}
Alec Radford, Jeffrey Wu, Rewon Child, David Luan, Dario Amodei, Ilya
  Sutskever, et~al.
\newblock Language models are unsupervised multitask learners.
\newblock {\em OpenAI blog}, 1(8):9, 2019.

\bibitem{song2015sun}
Shuran Song, Samuel~P Lichtenberg, and Jianxiong Xiao.
\newblock Sun rgb-d: A rgb-d scene understanding benchmark suite.
\newblock In {\em Proceedings of the IEEE conference on computer vision and
  pattern recognition}, pages 567--576, 2015.

\bibitem{sun2020scalability}
Pei Sun, Henrik Kretzschmar, Xerxes Dotiwalla, Aurelien Chouard, Vijaysai
  Patnaik, Paul Tsui, James Guo, Yin Zhou, Yuning Chai, Benjamin Caine, et~al.
\newblock Scalability in perception for autonomous driving: Waymo open dataset.
\newblock In {\em Proceedings of the IEEE/CVF conference on computer vision and
  pattern recognition}, pages 2446--2454, 2020.

\bibitem{tong2022videomae}
Zhan Tong, Yibing Song, Jue Wang, and Limin Wang.
\newblock Video{MAE}: Masked autoencoders are data-efficient learners for
  self-supervised video pre-training.
\newblock In {\em Advances in Neural Information Processing Systems}, 2022.

\bibitem{uy2019revisiting}
Mikaela~Angelina Uy, Quang-Hieu Pham, Binh-Son Hua, Thanh Nguyen, and Sai-Kit
  Yeung.
\newblock Revisiting point cloud classification: A new benchmark dataset and
  classification model on real-world data.
\newblock In {\em Proceedings of the IEEE/CVF international conference on
  computer vision}, pages 1588--1597, 2019.

\bibitem{wu2021balanced}
Tong Wu, Liang Pan, Junzhe Zhang, Tai WANG, Ziwei Liu, and Dahua Lin.
\newblock Balanced chamfer distance as a comprehensive metric for point cloud
  completion.
\newblock In A. Beygelzimer, Y. Dauphin, P. Liang, and J.~Wortman Vaughan,
  editors, {\em Advances in Neural Information Processing Systems}, 2021.

\bibitem{wu20153d}
Zhirong Wu, Shuran Song, Aditya Khosla, Fisher Yu, Linguang Zhang, Xiaoou Tang,
  and Jianxiong Xiao.
\newblock 3d shapenets: A deep representation for volumetric shapes.
\newblock In {\em Proceedings of the IEEE conference on computer vision and
  pattern recognition}, pages 1912--1920, 2015.

\bibitem{xie2022simmim}
Zhenda Xie, Zheng Zhang, Yue Cao, Yutong Lin, Jianmin Bao, Zhuliang Yao, Qi
  Dai, and Han Hu.
\newblock Simmim: A simple framework for masked image modeling.
\newblock In {\em Proceedings of the IEEE/CVF Conference on Computer Vision and
  Pattern Recognition}, pages 9653--9663, 2022.

\bibitem{Yu_2022_CVPR}
Xumin Yu, Lulu Tang, Yongming Rao, Tiejun Huang, Jie Zhou, and Jiwen Lu.
\newblock Point-bert: Pre-training 3d point cloud transformers with masked
  point modeling.
\newblock In {\em Proceedings of the IEEE/CVF Conference on Computer Vision and
  Pattern Recognition (CVPR)}, pages 19313--19322, June 2022.

\bibitem{zeng2022lift}
Yihan Zeng, Da Zhang, Chunwei Wang, Zhenwei Miao, Ting Liu, Xin Zhan, Dayang
  Hao, and Chao Ma.
\newblock Lift: Learning 4d lidar image fusion transformer for 3d object
  detection.
\newblock In {\em Proceedings of the IEEE/CVF Conference on Computer Vision and
  Pattern Recognition}, pages 17172--17181, 2022.

\bibitem{pvt}
Cheng Zhang, Haocheng Wan, Xinyi Shen, and Zizhao Wu.
\newblock Pvt: Point-voxel transformer for point cloud learning.
\newblock {\em International Journal of Intelligent Systems}, pages 1--24,
  2022.

\bibitem{zhang2022point}
Renrui Zhang, Ziyu Guo, Peng Gao, Rongyao Fang, Bin Zhao, Dong Wang, Yu Qiao,
  and Hongsheng Li.
\newblock Point-m2ae: Multi-scale masked autoencoders for hierarchical point
  cloud pre-training.
\newblock In {\em Advances in Neural Information Processing Systems}, 2022.

\bibitem{zhou2021ibot}
Jinghao Zhou, Chen Wei, Huiyu Wang, Wei Shen, Cihang Xie, Alan Yuille, and Tao
  Kong.
\newblock Image {BERT} pre-training with online tokenizer.
\newblock In {\em International Conference on Learning Representations}, 2022.

\bibitem{zhou2020end}
Yin Zhou, Pei Sun, Yu Zhang, Dragomir Anguelov, Jiyang Gao, Tom Ouyang, James
  Guo, Jiquan Ngiam, and Vijay Vasudevan.
\newblock End-to-end multi-view fusion for 3d object detection in lidar point
  clouds.
\newblock In {\em Conference on Robot Learning}, pages 923--932. PMLR, 2020.

\end{thebibliography}
}

\newpage
\appendix
\section*{Supplementary material}

\section{Baseline hyperparameters}
\label{appendix:baseline_params}
In this section, we present hyperparameters for the 3D OD model. For training the detector, we use the same loss functions as in the original SST implementation \cite{Fan_2022_CVPR}, but modify hyperparameters for the nuScenes dataset. Unless stated otherwise, the same set of parameters are used for pre-training, e.g., the voxelization parameters in Table \ref{tab:params_voxel}, the voxel encoder in Table \ref{tab:params_voxel_encoder}, and the SST encoder in Table \ref{tab:params_sst_encoder}. Table \ref{tab:params_detectionhead} specifies parameters used for downstream task training only.  

\begin{table}[ht]
    \centering
    \begin{tabular}{c|c}
        \toprule
        Parameter & Value  \\ \midrule
        Voxel size (m) & $0.5\times0.5\times8$ \\
        Max \#point & $\infty$ \\
        Max \#points/voxel & $\infty$ \\
        Max \#voxels & $\infty$ \\
        Point cloud range - $x$ & [-50 m, 50 m] \\
        Point cloud range - $y$ & [-50 m, 50 m] \\
        Point cloud range - $z$ & [-3 m, 5 m] \\
        Voxel grid shape (x,y,z) & (200,200,1)
        \\ \bottomrule
    \end{tabular}
    \caption{Parameters used for voxelization.}
    \label{tab:params_voxel}
\end{table}

\begin{table}[ht]
    \centering
    \begin{tabular}{c|c}
        \toprule
        Parameter & Value  \\ \midrule
        First linear layer & 64 output channels \\
        Second linear layer & 128 output channels
        \\ \bottomrule
    \end{tabular}
    \caption{Parameters used for voxel encoder.}
    \label{tab:params_voxel_encoder}
\end{table}

\begin{table}[ht]
    \centering
    \begin{tabular}{c|c}
        \toprule
        Parameter & Value  \\ \midrule
        Window size & $16\times16$ \\
        Padding levels (train) & [30, 60, 100, 200, 250] \\
        Padding levels (test) & [30, 60, 100, 200, 256] \\
        \#Layers & 8 \\
        Input dimension & 128 \\
        FFN hidden dimension & 256 \\
        \#Heads & 8 \\
        \#Attached conv. layers & 3 \\
        Conv. kernel size & $3\times3$ \\
        Conv. stride & 1 \\
        Conv. padding (per layer) & (1,1,2) \\
        Conv. in/out channels & 128\\
        Linear projection dim & 384 
        \\ \bottomrule
    \end{tabular}
    \caption{Parameters used for SST encoder. Note that the convolution layers are not used during pre-training. The padding levels refer to the grouping of windows when passing them through an encoder block, which allows for more efficient computations.}
    \label{tab:params_sst_encoder}
\end{table}

\begin{table}[ht]
    \centering
    \begin{tabular}{c|c}
        \toprule
        Parameter & Value  \\ \midrule
        \cellcolor[gray]{0.9} Class loss & \cellcolor[gray]{0.9} FocalLoss($\gamma = 2.0$,$\alpha= 0.25$) \\ 
        Bounding box loss & SmoothL1Loss($\beta = 1/9$)\\ 
        \cellcolor[gray]{0.9} Direction loss & \cellcolor[gray]{0.9} CrossEntropyLoss \\ 
        \multirowcell{3}{Bounding box \\ target weight \\ $x, y, z, w, l, h, \theta, v_x , v_y$} & \\ 
        & 1,1,1,1,1,1,1,0.1,0.1\\
        & \\ 
        \cellcolor[gray]{0.9} IOU class  &  \cellcolor[gray]{0.9}0.6\\
        \cellcolor[gray]{0.9} assignment threshold & \cellcolor[gray]{0.9}   \\ 
        \multirowcell{2}{IOU background \\ assignment threshold} & 0.3 \\
        &  \\ 
        \cellcolor[gray]{0.9} max NMS evaluations & \cellcolor[gray]{0.9} 1,000 \\ 
        NMS IOU threshold &  0.2\\ 
        \cellcolor[gray]{0.9} score threshold & \cellcolor[gray]{0.9} 0.05 \\ 
        min bbox size & 0 \\ 
        \cellcolor[gray]{0.9} max NMS predictions & \cellcolor[gray]{0.9} 500 \\ 
        ($\beta_{loc},\beta_{cls},\beta_{dir}$) & (1,1,0.2)
        \\ \bottomrule
    \end{tabular}
    \caption{Parameters used for the detection head. NMS stands for non-maximum suppression and is used during evaluation to filter predictions.}
    \label{tab:params_detectionhead}
\end{table}

\section{Pre-training hyperparameters} 
\label{appendix:pretrain_params}
In this section we present hyperparameters used for the decoder and reconstruction head using during pre-training, see Tables \ref{tab:params_voxel_decoder} and \ref{tab:params_voxel_reconstructionhead}.
\begin{table}[ht]
    \centering
    \begin{tabular}{c|c}
        \toprule
        Parameter & Value  \\ \midrule
        Window size & $16\times16$ \\
        Padding levels (train) & [30, 60, 100, 200, 250] \\
        Padding levels (test) & [30, 60, 100, 200, 256] \\
        \#Blocks & 8 \\
        Input dimension & 128 \\
        FFN hidden dimension & 256 \\
        \#Heads & 8 \\
        \#Empty voxels & $0.1\cdot$\#voxels
        \\ \bottomrule
    \end{tabular}
    \caption{Parameters for the decoder used during pre-training. The padding levels refer to the grouping of windows when passing them through a decoder block, which allows for more efficient computations.}
    \label{tab:params_voxel_decoder}
\end{table}

\begin{table}[ht]
    \centering
    \begin{tabular}{c|c}
        \toprule
        Parameter & Value  \\ \midrule
        Empty voxel loss & BinaryCrossEntropy
        \\
        Number of points loss & SmoothL1($\beta=1$) \\
        \#Predicted points (Chamfer) & 10 \\
        \#Max GT points (Chamfer) & 100 \\
        $\alpha_c$ & 1 \\
        $\alpha_{np}$ & 1 \\
        $\alpha_{occ}$ & 1
        \\ \bottomrule
    \end{tabular}
    \caption{Parameters for the reconstruction head used during pre-training.}
    \label{tab:params_voxel_reconstructionhead}
\end{table}

\section{Results with two sweeps} 
\label{appendix:two_sweeps}
\begin{table}[ht]
    \centering
    \begin{tabular}{cccc}
        \toprule
        Voxel size (m) & Encoder depth & mAP & NDS  \\ \midrule
        0.25 & 6 & 31.43 & 50.76 \\
        0.30 & 6 & 35.93 & 52.60 \\
        0.50 & 6 & 42.79 & \textbf{55.54} \\
        0.50 & 8 & \textbf{43.60} & 55.19 \\
        0.70 & 6 & 41.73 & 54.69 \\
        0.70 & 8 & 31.31 & 54.21 
        \\ \bottomrule
    \end{tabular}
    \caption{Performance on the nuScenes validation dataset for a model using 2 sweeps and without any pre-training.}
    \label{tab:encoder_and_voxel_size_ablation}
\end{table}
\subsection{Data efficiency}
\begin{table*}[htb]
    \centering
    \begin{tabular}{c|c|cc|cccccccc}
        \toprule
        Dataset fraction & Pre-trained & mAP & NDS & ped. & car & truck & bus & barrier & T.C. & trailer & moto. \\
        \midrule
        \multirow{2}{*}{0.2} & \xmark & 35.54 & 47.79 & 62.5 & 73.6 & 35.8 & 41.7 & 49.4 & 31.2 & 14.8 & 29.5\\
         & \cmark & \textbf{39.95} & \textbf{51.60} & \textbf{69.1} & \textbf{75.7} & \textbf{40.4} & \textbf{48.1} & \textbf{54.6} & \textbf{39.3} & \textbf{18.2} & \textbf{33.6} \\
        \hdashline
        \multirow{2}{*}{0.4} & \xmark & 38.99 & 51.41 & 66.7 & 76.8 & 39.8 & 48.2 & 51.9 & 34.9 & 17.9 & 33.4 \\
         & \cmark & \textbf{43.15} & \textbf{54.46} & \textbf{71.4} & \textbf{77.5} & \textbf{43.0} & \textbf{54.2} & \textbf{57.5} & \textbf{41.0} & \textbf{20.7} & \textbf{38.4} \\
        \hdashline
        \multirow{2}{*}{0.6} & \xmark & 41.29 & 53.28 & 69.1 & 77.4 & 41.3 & 50.8 & 54.9 & 37.4 & 20.0 & 38.7 \\
         & \cmark & \textbf{43.77} & \textbf{55.29} & \textbf{72.1} & \textbf{77.8} & \textbf{44.0} & \textbf{54.5} & \textbf{57.4} & \textbf{43.8} & \textbf{21.8} & \textbf{40.7} \\
        \hdashline
        \multirow{2}{*}{0.8} & \xmark & 42.26 & 54.24 & 69.2 & \textbf{77.8} & 40.8 & 53.6 & 55.4 & 40.2 & 19.4 & 39.6 \\
        & \cmark & \textbf{43.96} & \textbf{55.57} & \textbf{72.2} & \textbf{77.8} & \textbf{43.0} & \textbf{55.0} & \textbf{57.9} & \textbf{42.6} & \textbf{22.1} & \textbf{41.7} \\
        \hdashline
        \multirow{2}{*}{1.0} & \xmark & 43.60 & 55.19 & 69.9 & \textbf{78.9} & \textbf{43.1} & \textbf{55.7} & 56.7 & 39.5 & \textbf{21.4} & 41.5 \\
         & \cmark & \textbf{44.62} & \textbf{56.00} & \textbf{72.1} & {78.3} & {43.0} & {54.5} & \textbf{57.3} & \textbf{43.2} & {21.2} & \textbf{44.9} \\
        \bottomrule
    \end{tabular}
    \caption{mAP, NDS, and AP per class on the nuScenes validation data for pre-trained and randomly initialized models when varying the amount of \textit{labeled} data. Pre-training and fine-tuning is done with \textit{two} aggregated point cloud sweeps without intensity information. ped.=pedestrian. T.C.=traffic cone. moto.=motorcycle.}
    \label{tab:data_efficiency_2_sweeps}
\end{table*}

Table \ref{tab:data_efficiency_2_sweeps} shows the data efficiency results when pre-training and fine-tuning were done with two aggregated sweeps. Similar to the results for 10 sweeps in Table 1, we see that our Voxel-MAE brings a substantial performance increase compared to the fully supervised baseline. The baseline reaches 43.6 mAP and 55.19 NDS when using the entire training dataset. The model pre-trained with Voxel-MAE outperforms this baseline when using only 60\% of the annotated data with 43.77 mAP and 55.29 NDS. 

Same as for the experiments with 10 sweeps the pre-trained models consistently improve upon their baseline in terms of mAP and NDS regardless of dataset fraction. Further, the largest improvements can be found for models fine-tuned on 20\% of the annotations, indicating the effectiveness of our method when the amount of unlabeled data is large compared to the annotated one. However, also when using all available annotations, pre-training can increase detection performance. 

\subsection{Encoder depth and voxel size}
In Table \ref{tab:encoder_and_voxel_size_ablation} we study how baseline performance varies with different number of encoder layers and voxel size. For reference, the original SST model was tuned toward the Waymo Open dataset and used 6 encoder layers and a voxel size of $0.32\times0.32\times6$ m. However, for nuScenes, we found better performance with 8 encoder layers and a voxel size of $0.5\times0.5\times8$ m.

\end{document}